# Multi-attention-based soft partition network for vehicle re-identification

Sangrok Lee[1], Taekang Woo[2] and Sang Hun Lee[3,*]

[1]Vision Intelligence for Transportation and Automobiles Laboratory, MODULABS, Seoul 06252, South Korea
[2]Commerce AI, NAVER Corporation, Seongnam 13561, South Korea
[3]Graduate School of Automotive Engineering, Kookmin University, Seoul 02707, South Korea
*Corresponding author. E-mail: shlee@kookmin.ac.kr

## Abstract

Vehicle re-identification helps in distinguishing between images of the same and other vehicles. It is a challenging process because of significant intra-instance differences between identical vehicles from different views and subtle inter-instance differences between similar vehicles. To solve this issue, researchers have extracted view-aware or part-specific features via spatial attention mechanisms, which usually result in noisy attention maps or otherwise require expensive additional annotation for metadata, such as key points, to improve the quality. Meanwhile, based on the researchers' insights, various handcrafted multi-attention architectures for specific viewpoints or vehicle parts have been proposed. However, this approach does not guarantee that the number and nature of attention branches will be optimal for real-world re-identification tasks. To address these problems, we proposed a new vehicle re-identification network based on a multiple soft attention mechanism for capturing various discriminative regions from different viewpoints more efficiently. Furthermore, this model can significantly reduce the noise in spatial attention maps by devising a new method for creating an attention map for insignificant regions and then excluding it from generating the final result. We also combined a channel-wise attention mechanism with a spatial attention mechanism for the efficient selection of important semantic attributes for vehicle re-identification. Our experiments showed that our proposed model achieved a state-of-the-art performance among the attention-based methods without metadata and was comparable to the approaches using metadata for the VehicleID and VERI-Wild datasets.

## 1. Introduction

Vehicle re-identification (Re-ID) identifies the same vehicle from numerous images. It searches for the same car in gallery images as depicted in a given query image. This task received considerable attention recently because the Re-ID technology can be used to analyse traffic flow, which is used to build smart cities and is an essential technology for surveillance systems. Vehicle Re-ID is particularly challenging because their exteriors can be captured in numerous environments, and different lights and viewpoints can cause significant intra-instance differences. Other vehicles can also look similar because of matching colours and general vehicle types.

Recent studies have used convolutional neural networks (CNNs) and metric learning methods (Jiang *et al.*, 2018; Khorramshahi *et al.*, 2019a; Lee *et al.*, 2020, 2022; Liu *et al.*, 2018; Tang *et al.*, 2017; Zhang *et al.*, 2017; Zhou & Shao, 2018). In the latter, vehicle images are encoded to a representative vector in embedding space, and the distances between the vectors are compared. Thus, it is crucial to select robust features to accommodate variations in environments, light conditions, and viewpoints.

To effectively identify the same vehicles, several studies have adopted metadata attributes (e.g., orientation, colour, type, keypoint, viewpoint, and spatiotemporal information) (Jiang *et al.*, 2018; Shen *et al.*, 2017; Tang *et al.*, 2017; Wang *et al.*, 2017). Recent studies have semantically divided vehicles into its parts using image segmentation and mask them; furthermore, features are extracted from the segmented regions (Chen *et al.*, 2020b; He *et al.*, 2019; Meng *et al.*, 2020). These methods can compare not only global appearance but also vehicle parts; thus, subtle vehicle parts can be embedded and compared. However, they have a major drawback because they require extensive image annotation. Particularly, labelling vehicle parts, such as segmentation and bounding box creation, requires significant time than labelling images. According to the report (Lin *et al.*, 2014), segmenting takes 15 times longer than spotting object locations and 60 times longer than image labelling.

In recent years, the combination of deep learning and visual attention mechanism has been explored to increase the performance of vehicle Re-ID tasks (Wang *et al.*, 2019). The visual attention is formed by training deep neural networks to learn which areas need to be focused on in each new image. Based on their insights, many researchers have proposed various handcrafted spatial attention-based models equipped with multi-attention branches for specific viewpoints and vehicle parts. For instance, some researchers have developed multi-view attention networks (MVANs) in which each branch learns viewpoint-specific features for three to five views (e.g., front, rear, top, and side views) (Chen *et al.*, 2020a, b; Teng *et al.*, 2021; Zhang *et al.*, 2020; Zhou & Shao, 2018). Others have devised part-oriented attention networks that localize various salient vehicle parts, such as the windshield and car head, for hard part-level attention and then provide an additional soft attention refinement at the pixel level (Guo *et al.*, 2019; Zhang

**Received:** August 17, 2022. **Revised:** February 7, 2023. **Accepted:** February 15, 2023

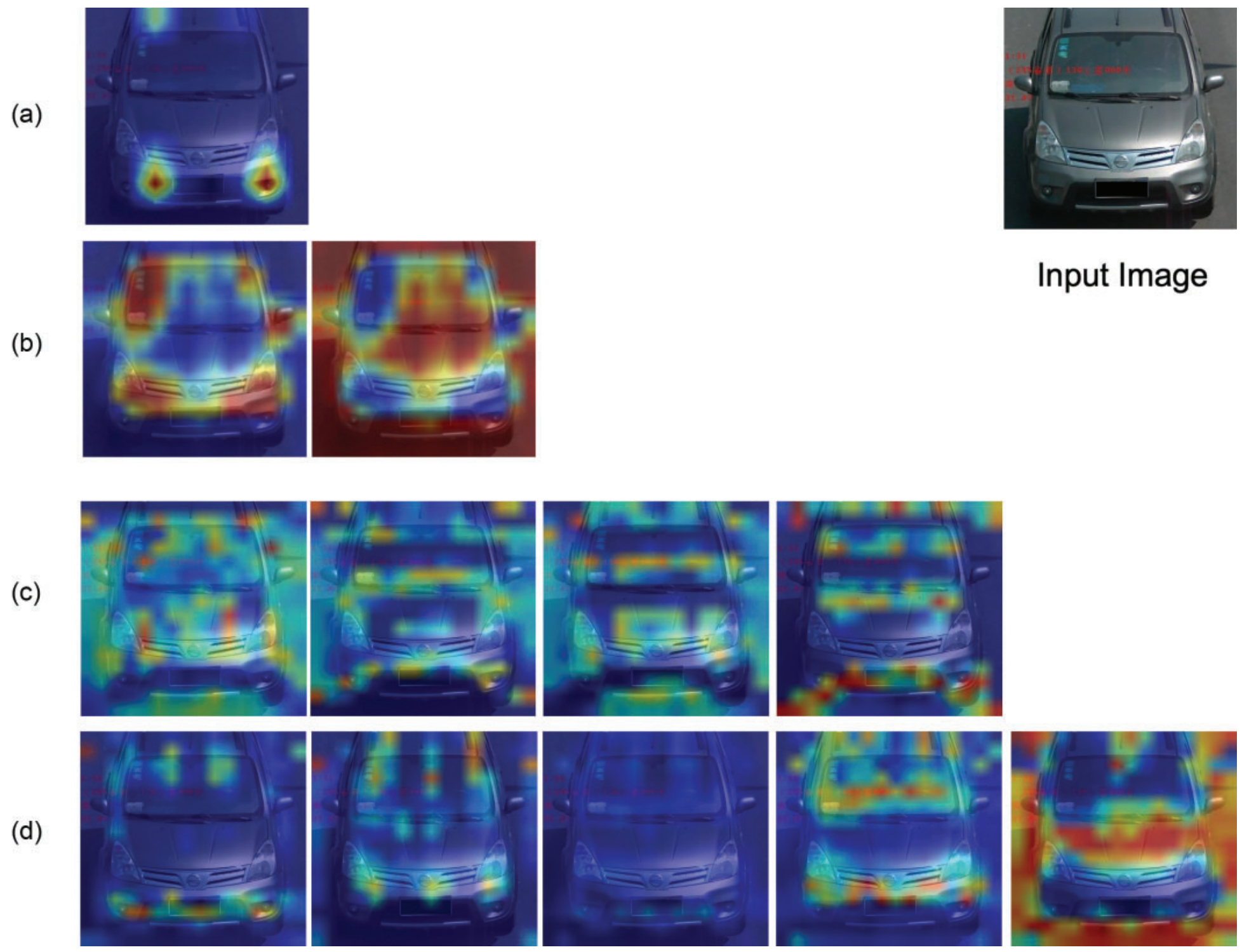

**Figure 1:** Activation heat maps, associated with attention weights, visualized by Grad-CAM (Zhang *et al.*, 2020) for (a) the baseline model with single attention, (b) the proposed model with double attentions, (c) the baseline model with four attentions, and (d) the proposed model with five attentions. The right-most heat maps in (b) and (d) represent the attention map for background or insignificant areas, and the associated feature map is discarded from producing the output results. Compared with the baseline, the proposed model suppresses noise from the spatial attention maps and captures subtle discriminative regions useful for distinguishing vehicles.

*et al.*, 2020). However, as shown in Figs. 1(a) and (c), spatial attention mechanisms are usually hampered by noise generated attention maps or require expensive additional annotation for metadata attributes, such as keypoints, to improve quality (Chen *et al.*, 2020a). Moreover, the handcrafted spatial attention architecture does not ensure that the number and types of attention branches are optimal for real-world Re-ID tasks.

To solve these problems, we proposed the multi-attention-based soft partition (MUSP) network, which uses multiple and soft attention mechanisms in the spatial and channel-wise directions, as shown in Fig. 2. Multi-attention enables the model to learn diverse features from various discriminated regions and different viewpoints without introducing any artificial part-specific and view-dependent attention branches. Soft attention gives continuous values of the region mask; its weights can be learned through backpropagation during training, as it is fully differentiable (Wang *et al.*, 2019). We also combined a channel-wise attention mechanism with a spatial attention mechanism to select the better semantic attributes that are meaningful for vehicle Re-ID. Finally, we introduced a novel method for removing noise from spatial attention maps. In our framework, as shown in Fig. 2, we first apply multiple spatial attention weights to feature maps encoded from a vehicle image, followed by channel-wise attention weights. Here, the last feature map with spatial attention applied is excluded from the input of the channel-wise attention branch. Thus, the edge from the node of the final feature map to the next step in the computation graph is dropped out. Then, the spatial attention weights of the indiscriminative regions are aggregated into the last attention map due to backpropagation training. Simultaneously, the spatial attention weights of the salient areas contributing to the final result are collected in the other attention maps. Figure 1(c) shows the heat map of the resulting attention weights for a four-attention-based model in which all spatial attention maps are used for training and inference. The attention maps contain a lot of noise. However, as shown in Fig. 1(d), our model's first to fourth attention maps are clean. Here, the last fifth map, which is excluded from further computation in the pipeline, contains attention weights that are highly activated to the background or irrelevant parts. Our experiments revealed that our proposed model achieved a state-of-the-art (SOTA) performance among the attention-based methods without metadata and was comparable to the approaches using metadata for the VehicleID (Liu *et al.*, 2016c) and VERI-Wild (Liu *et al.*, 2016a) datasets.

In summary, the main contributions of our approach are as follows:

- (i) We proposed a novel MUSP network for vehicle Re-ID. With multiple spatial attention and channel-wise attention mechanisms, the model can learn to highlight the most discriminant regions and suppress the distraction of irrelevant parts without introducing any artificial part-specific and view-dependent attention branches.
- (ii) We developed a novel method that reduces noise in spatial attention maps so significantly that we solved the problem of spatial attention mechanisms generating noise in attention maps.
- (iii) Our approach achieved SOTA performance among the attention-based approaches that did not use metadata and

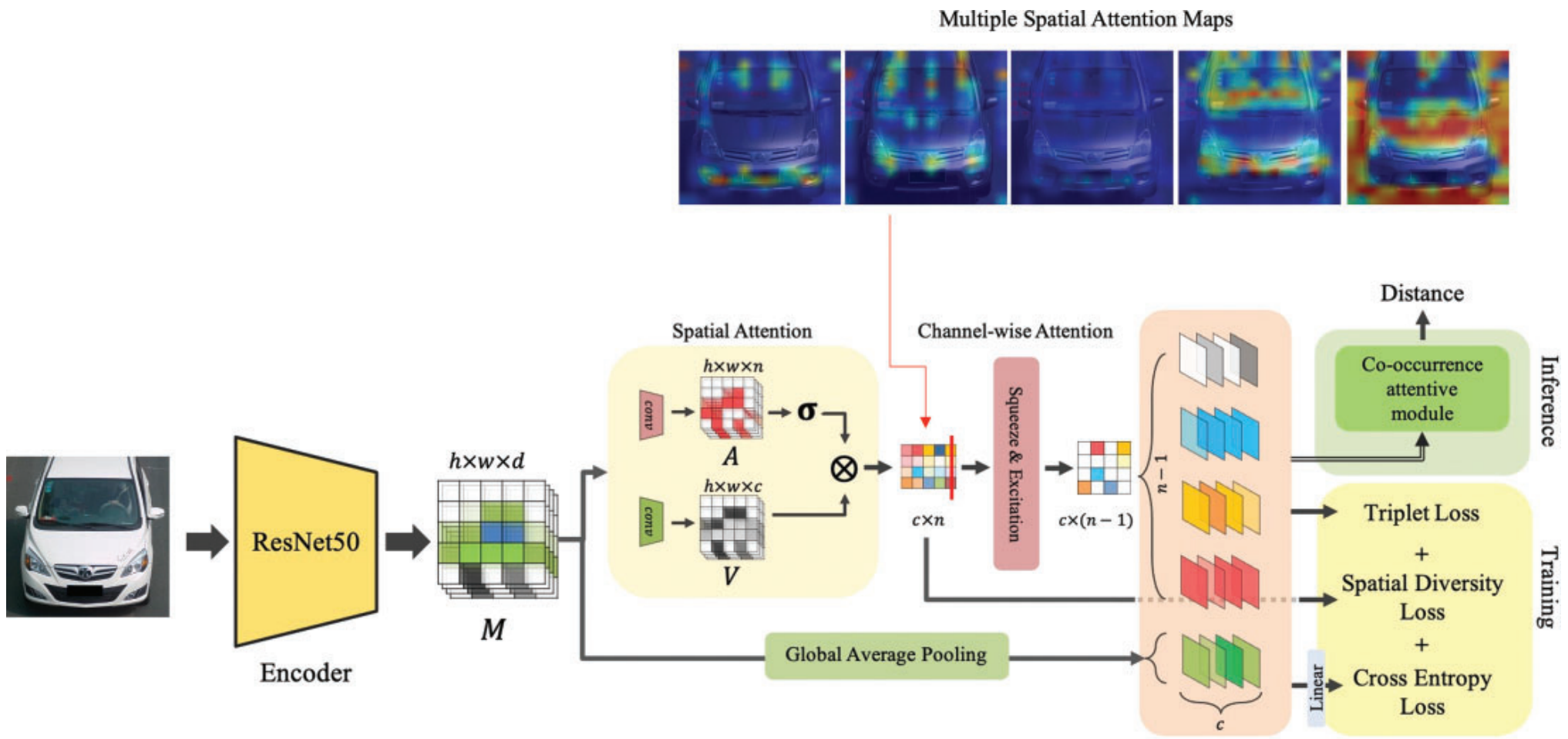

**Figure 2:** Proposed MUSP pipeline. First, the image is placed in the backbone network (ResNet50). The features extracted from the backbone network go through the attention-based network, which is sequentially composed of spatial and channel-based attention modules. The $n - 1$ vectors obtained from the attention-based network and a global vector obtained using average pooling are used to train the model using triplet, spatial diversity, and cross-entropy losses. In the inference stage, a total of $n$ vectors are calculated using the co-occurrence attentive module.

showed comparable performance to the methods that used metadata in the experiments using the VehicleID (Liu *et al.*, 2016c) and VERI-Wild (Liu *et al.*, 2016a) datasets.

The rest of the paper is organized as follows: Section 2 reviews recent related studies in detail, particularly on attention-based approaches. Section 3 describes the architecture of the proposed system and detailed algorithms for each step. Section 4 discusses loss functions for training the model. Section 5 demonstrates our approach's effectiveness and competitiveness on numerous challenging vehicle Re-ID benchmarks through extensive experiments. Finally, Section 5 concludes this paper.

## 2. Related Work

Vehicle Re-ID technology has advanced enormously and is strongly driven by access to several large datasets (Kanaci *et al.*, 2018; Liu *et al.*, 2016c; Lou *et al.*, 2019b), which enable models to be trained and tested more closely in real-world environments. Deep and metric learnings have been used for vehicle Re-ID tasks. Additional representative features must be extracted when embedding vehicle images in the feature space to increase metric learning performance. Consequently, numerous attempts to use the metadata of vehicles have been introduced, such as orientation, colour, type, key points, viewpoint, and spatiotemporal data. Additionally, most researchers have combined deep learning and visual attention mechanisms to extract the features of the distinguishing regions to improve the accuracy of the vehicle Re-ID task. The problem here is to identify the type of attention mechanism that is more efficient and adaptive among the various types of attention networks. Recently, various methods of using generative adversarial networks (GANs) have also been proposed (Zhou & Shao, 2017, 2018); however, there is a large gap between the generated features and reality due to the limitations of the generation ability of the existing GANs and the lack of adversarial samples.

### 2.1 Metadata

Temporal data have been adopted by several studies for efficient vehicle Re-ID (Jiang *et al.*, 2018; Liu *et al.*, 2016c; Shen *et al.*, 2017). Shen *et al.* (2017) used temporal information to track gradual vehicle changes from different cameras, which enabled them to recognize the same vehicle that looked different and overcome the limitations of the method using only spatial information. However, this method has a disadvantage because a continuous stream of images is required. Liu *et al.* (2016c) re-ranked the images using temporal information after vehicle detection. This approach requires the temporal information of each vehicle even in the inference stage. Furthermore, Jiang *et al.* (2018) used temporal information and spatial information for re-ranking.

Previous studies used vehicle key points for vehicle Re-ID (Khorramshahi *et al.*, 2019a; Wang *et al.*, 2017). Wang *et al.* (2017) not only used temporal information but also estimated orientation using key points and extracted orientation-invariant features to improve the performance of vehicle Re-ID. Khorramshahi *et al.* (2019a) used key points to exploit local features. The key-point-based method has the following disadvantages: it cannot cope with various types of vehicles that do not exist in the training data and additional key point labels are required.

Recent studies (Chen *et al.*, 2020b; Liu *et al.*, 2018; Meng *et al.*, 2020) introduced a method of segmenting and comparing vehicle parts using metadata. This method is similar to how humans identify objects by segmenting the parts of a vehicle and comparing each part separately. Liu *et al.* (2018) used a detection model to segment vehicle parts. Chen *et al.* (2020b) proposed a model that segments vehicle parts in a weakly supervised method using the orientation of the vehicle to improve the performance of vehicle Re-ID. Meng *et al.* (2020) used a supervised segmentation model to divide vehicles. These methods have improved performance but have the disadvantage of requiring additional annotations or models. Detection and segmentation require numerous data resources, and the model is heavy.

### 2.2 Visual attention

The visual attention mechanism enables deep neural networks to learn the areas that must be focused during training. Figure 1 shows the learnt attention maps of several vehicle images, and the highlighted areas correspond to the subtle and discriminative regions. The attention mechanism automatically extracts the distinct features of the regions, thereby resulting in improved accuracy in the vehicle Re-ID.

**Table 1.** Attention-based approaches for vehicle Re-ID.

| Method | Branch count | Attention branch | Spatial/ channel | Soft/ hard | Extra annotation | Loss | Backbone |
|---|---|---|---|---|---|---|---|
| VAMI (Zhou & Shao, 2018) | 5 | Five views | Sp | S | Viewpoint, model, type, and colour | Id, adversarial, model, type, and colour | F-Net |
| SCAN (Teng et al., 2018) | 2 | Spatial and channel | Sp and Ch | S | | Triplet and id | VGG-CNN-M-1024 |
| TAMR (Guo et al., 2019) | 3 | Global, windscreen, and car-head | Sp | S | Model | Ranking(2), pairwise, and id | ResNet18 |
| VCAM (Chen et al., 2020a) | 1 | View-channel | Ch | S, H+S | Viewpoint | Viewpoint, triplet, and id | ResneXt-101 |
| SPAN (Chen et al., 2020b) | 3 | Front, side, and rear | Sp | S | Viewpoint | Triplet, id, mask reconstruction, area constraint, and spatial diversity | ResNet50 |
| SAVER (Khorramshahi et al., 2020) | 1 | VAE-residual | Sp | S | | Triplet, id, and reconstruction | ResNet50 |
| PVEN (Meng et al., 2020) | 4 | Multi-view (front, side, top, and back) | Sp | H | View segmentation | Triplet and id | ResNet50 |
| PGAN (Zhang et al., 2020) | 16 | Parts | Sp | H+S | Part | Triplet and id | ResNet50 |
| MSA (Zheng et al., 2020) | 3 | Multi-scale | Sp and Ch | S | | Id | ResNet50 |
| MVAN (Teng et al., 2021) | 6 | {Front,side,rear} × {global,local} | Sp | S | Viewpoint | Triplet and id | VGG18 and ResNet50 |
| PSA (Yang et al., 2021) | 1 | Global | Sp | S | | Triplet and id | ResNet50 |
| SSIA (Li et al., 2021) | 1 | Global | Sp and Ch | S | Rotation | Triplet, id and rotation | ResNet50 |
| CAL (Rao et al., 2021) | 16 | Real and counterfactual | Sp | S | | Triplet and id | ResNet50 |
| MUSP (This work) | 10 | Multi-map | Sp and Ch | S | | Triplet, id, and spatial diversity | ResNet50 |

Representative attention-based models are shown in Table 1. Most models have multiple attention branches. Viewpoint and vehicle parts are widely used as criteria for creating branches. Typical attention-based models comprise a trunk branch to learn a global feature representation and multiple part- or viewpoint-specific branches to learn feature representations dependent on parts or viewpoints (Chen *et al.*, 2020b; Guo *et al.*, 2019; Zhang *et al.*, 2020).

Most models have spatial attention, whereas others have channel-wise attention. Since a channel-wise feature map is essentially for detecting the corresponding semantic attributes, channel-wise attention can be regarded as the process of selecting semantic attributes that are meaningful or potentially helpful for achieving the goal. Therefore, channel-wise feature maps are usually used to detect discriminative vehicle parts, such as the windshield or tires, while spatial attention feature maps are used to extract viewpoint-aware features (Chen *et al.*, 2020a).

Soft attention uses soft shading to focus on regions, whereas hard attention uses image cropping to focus on regions. Moreover, soft attention can be learnt using gradient descent, whereas hard attention cannot be trained because there is no derivative for the procedure ‘crop the image here’. All models use soft attention by default, and hard attention is used when the model has vehicle part branches.

Many models attempt to improve performance using metadata, which may require additional annotations. The triplet and cross-entropy losses are used for basic training, and if any label is used, its cross-entropy loss is added. The most used backbone network is ResNet50, which is also used for our model.

Table 1 shows that recent attention-based approaches usually adopted multiple attention branches for viewpoints or discriminative parts and introduced spatial and channel-wise attention modules, which are connected in serial or parallel depending on the developer’s choice. However, there are still opportunities to improve existing approaches, as the number of attention branches, associated viewpoints, and parts and metadata usage are hyperparameters or the developer’s choice.

We now classify and describe the models listed in Table 1 into three groups: the viewpoint-centric, part-centric, and miscellaneous attention models.

### 2.2.1 *Viewpoint-centric attention models*

Zhou and Shao (2018) proposed a viewpoint-aware attentive multi-view inference (VAMI) model and an adversarial training architecture to produce attention maps for different viewpoints. They further generated multi-view features from a single-view input image. However, the VAMI requires the viewpoint and attribute information of the vehicles to train the network for the single-view feature extraction of the vehicle. Additionally, because of the lack of direct supervision on the generated attention maps, the attention outcomes are noisy and would unfavourably affect the learning of the network.

Chen *et al.* (2020a) proposed the viewpoint-aware channel-wise attention mechanism (VCAM), which uses the viewpoint of a captured image to generate an attentive weight. They assumed that the viewpoint of the image determines the visibility of the vehicle parts and that channel-wise attentive weight is related to the visibility of the corresponding part. The Re-ID feature extraction module is incorporated to the channel-wise attention mech-anism to extract the viewpoint-aware feature for Re-ID matching. However, this method requires viewpoint annotations on vehicle images for supervised learning of the viewpoint estimation layer.

Meng *et al.* (2020) proposed a parsing-based view-aware embedding network (PVEN) to achieve view-aware feature alignment and enhance vehicle Re-ID. A network parses a vehicle into four different views and then aligns the features using mask average pooling to provide a fine-grained representation of the vehicle. Furthermore, common-visible attention was designed to focus on the common-visible views, which not only shortens the distance among intra-instances but also enlarges their discrepancy. Thus, PVEN can capture the stable and discriminative information of the same vehicle and outperforms the previous methods by a large margin. However, it requires additional annotation for training the vehicle part parser, which consumes considerable labour time.

Chen *et al.* (2020b) proposed a dedicated semantics-guided part-attention network (SPAN) that generates attention masks for three different vehicle views (i.e., front, side, and rear). The part-attention masks enable the network to extract separately discriminative features in each part (or view). Moreover, to recognize correctly two positive images, the features of the co-occurrence view are emphasized when evaluating the feature distance of two images. However, to train the part-attention network, it is crucial to provide viewpoint semantic labels although they are not pixel-level but image-level labels.

Teng *et al.* (2021) proposed a MVAN, where each branch learns a viewpoint-specific feature for the front, rear, and side views. Furthermore, a spatial attention model is introduced into each branch to learn specific local cues for different viewpoints. The viewpoint-specific features may outperform the general features learned using a uniform network because they can focus on a limited range of views. However, to train the model to estimate the viewpoint of an input vehicle image, we must annotate the viewpoints on the image dataset.

### 2.2.2 *Part-centred attention models*

Guo *et al.* (2019) proposed a two-level attention network supervised using a multi-grain ranking loss (TAMR) to learn efficient feature embeddings for vehicle Re-ID tasks. The two-level attention network consists of the hard part-level and soft pixel-level attention. The former is designed to localize the salient vehicle parts, such as the windscreen and car head. The latter provides an additional attention refinement at a pixel-level to focus on the distinctive characteristics within each part. The model consists of two groups of three branches: (i) a trunk branch for learning global feature representation with soft attention modules and (ii) two salient part branches with two-level hard and soft attention modules. Additionally, they proposed a coarse-to-fine multi-grain ranking loss to further enhance the discriminative ability of the learned features. However, as the windscreen and car head are only visible from the front, the branches of the parts do not provide useful information from the rear or side views.

Zhang *et al.* (2020) introduced a part-guided attention network (PGAN) that combines part-guided bottom-up and top-down attention, global features, and partial visual features in an end-to-end framework. First, the PGAN detects the locations of the various parts and salient regions and generates hard attention to them. Then, soft attention weights are learned for candidate parts to highlight the most discriminative regions and suppress the distraction of irrelevant parts. Finally, PGAN aggregates the global appearance and local features to improve the feature performance. However, we must annotate vehicle images to bounding boxes and class labels to train an object detector to recognise vehicle parts or regions.

### 2.2.3 *Miscellaneous attention models*

Teng *et al.* (2018) developed a spatial and channel attention network (SCAN), which embeds spatial and channel attention branches behind convolutional layers to highlight the outputs in discriminative regions and channels, respectively. The attention branches and convolutional layers are jointly trained using triplet and cross-entropy losses. However, since this method only focuses on the global feature map regardless of viewpoints, viewpoint-centred or part-focused local attention is crucial.

Khorramshahi *et al.* (2020) proposed the self-supervised attention for vehicle re-identification (SAVER), which generates an attention map using a variational auto-encoder (VAE). First, the SAVER generates a coarse reconstruction image; then, it performs its pixel-wise difference from the original image to construct a residual image. The residual image contains crucial details required for Re-ID and acts as a pseudo-saliency or pseudo-attention map that highlights discriminative regions in the image. Finally, the convex combination (with trainable parameter $\alpha$) of the reconstructed and residual image is calculated and passed through the Re-ID backbone for deep feature extraction. The SAVER benefits from self-supervised attention generation and eliminates the requirement of having extra annotations for key points of vehicles and bounding boxes of the parts, which are used to train specialized detectors. However, it only learns instance-specific discriminative features but ignores significant viewpoint changes between images of identical vehicles.

Zheng *et al.* (2020) proposed a multi-scale attention framework (MSA), which includes multi-branch subnetworks that generate different scales of feature maps using bilinear interpolation. Each subnetwork has a spatial-channel attention block, as shown in the SCAN. The multi-scale mechanism on the feature map level helps supply the missing information caused by pooling operations. However, this framework does not consider view-dependent attention, which is crucial to the vehicle Re-ID.

Li *et al.* (2021) first introduced self-supervised representation learning to discover geometric features and applied an interpretable attention module to condense these features. Unlike the other attention model, which is automatically learned using the backpropagation algorithm, the proposed attention model takes the local spatial maxima and channel-wise maxima to construct an attention map. By sharing the encoder of the self-supervised learning module with the attention module, the model can discover discriminative features from automatically found geometric locations without corresponding supervision. The discovered locations are mainly the corners of the body of a vehicle.

Rao *et al.* (2021) presented the counterfactual attention learning (CAL) method based on causal inference to learn attention that is more effective. They designed a framework to quantify the quality of attention by comparing the effects of facts and counterfactuals on the final prediction. Furthermore, they proposed to maximize the difference to encourage the network to learn visual attention that is more effective.

## 3. Proposed Method

The proposed multi-attention network comprises a backbone network, which is used to encode a convolutional feature map for a given image, and an attention-based network used to extract a set of weighted feature vectors, each of which focuses on a specific vehicle region. The latter consists of two modules: the spatial attention module, for soft partitioning of vehicle regions, and the channel-wise attention module, which is based on the squeeze and excitation (SE) method (Hu *et al.*, 2020). The weighted feature vectors are used to compare the distance between images for metric learning and are fed to a classifier to predict vehicle ID. The classifier includes batch normalization (Ioffe & Szegedy, 2015) and linear layers (Luo *et al.*, 2019). $n-1$ classifiers are applied to $n-1$ weighted feature vectors, excluding the background vector. Figure 2 depicts the overall architecture of the MUSP, and its components are described in detail in the rest of this section.

## 3.1 Feature extraction from vehicle images

We selected ResNet50 (He *et al.*, 2016) as a backbone for feature extraction, removed the last fully connected (FC) layer, and used the output of the last convolution layer. Thus, the feature extraction process is as follows:

$$\mathbf{M} = CNN_{\mathrm{B}}(\mathbf{I}), \mathbf{M} \in \mathbb{R}^{h\times w\times d}, \tag{1}$$

where $CNN_{\mathrm{B}}$ is the base network, $\mathbf{M}$ is a feature map extracted from $\mathbf{I}$, and $h$, $w$, and $d$ are dimensions of $\mathbf{M}$, which depend on the feature extractor and input image $\mathbf{I}$.

## 3.2 Spatial multi-attention mechanism

We use vehicle partitioning to extract subtle vehicle parts for vehicle Re-ID (Chen *et al.*, 2020b; Meng *et al.*, 2020; Sun *et al.*, 2018) and an attention method to refine the embedded features. Khorramshahi *et al.* (2019b) proposed a method for detecting and recropping a vehicle during pre-processing to reduce background regions. They used a detection model and bounding box annotation to depress the noisy background. We assume that the same function can be processed within the deep learning model without additional models or artificial intervention. Figure 1(b) illustrates that the vehicle area was accurately recognized without additional annotation or detection. Meng *et al.* (2020) found that subtle vehicle components significantly impact the division of parts. However, they cannot be captured accurately using single attention because attention focuses on easily comparable features, such as headlights and bumpers. Therefore, we used spatially separated multiple attention to focus on different vehicle areas. This distributed attention can consider different parts; thus, the model can see and compare more vehicle details. Consequently, we designed a spatial multiple attention mechanism.

We apply convolution layers to feature $\mathbf{M}$, which is encoded using the backbone network, to extract two feature maps for attention weights ($\mathbf{A}$) and values ($\mathbf{V}$). An attention feature map has $n$ channels with size $h \times w$. Each channel corresponds to a vehicle part. A value feature map has $c$ channels with size $h \times w$. Attention weights are normalized by the softmax function and are multiplied by the corresponding values to obtain $n$ weighted values, to which average pooling is applied to extract final weighted feature vectors $\{\mathbf{f}_i\}_{i=0,n}$. Note that the softmax function is applied to the last dimension $n$, rather than the spatial dimension $hw$. The attention weight has exclusive activation at each spatial point of the value map. We discard the final weighted feature vector $\mathbf{f}_n$. However, if the final vector is not used in the subsequent layers of the network, only the rest vectors are trained to pay attention to the discriminative regions. Therefore, the attention for the indiscriminative regions, such as the background, is assigned to the discarded vector, which is noise (Fig. 3). The entire process can be summarized as follows:

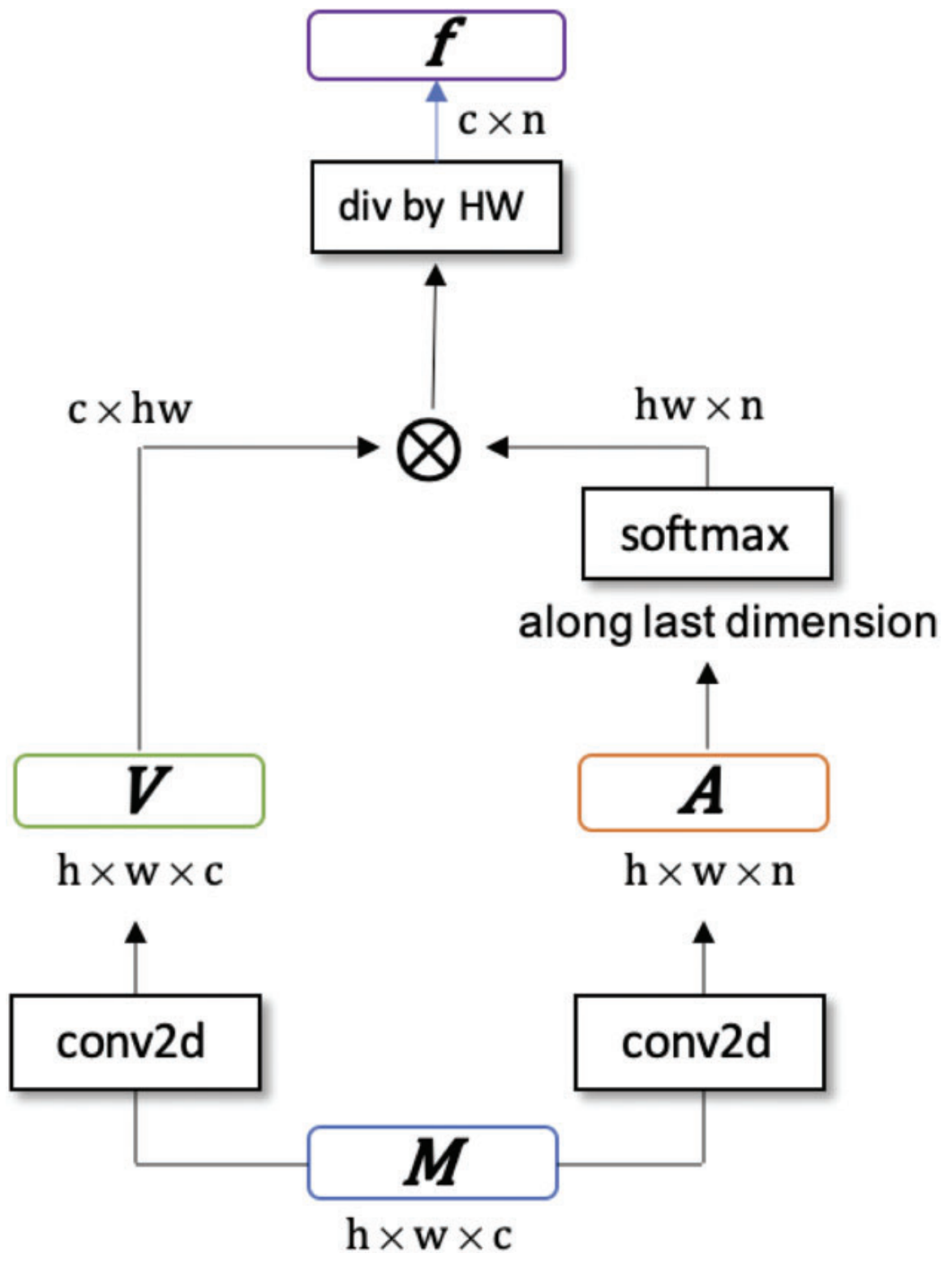

**Figure 3:** Proposed spatial attention implementation, where ⊗ denotes matrix multiplication.

$$\mathbf{V} = CNN_{\mathrm{VB}}(\mathbf{M}), \mathbf{V} \in \mathbb{R}^{h\times w\times c}, \tag{2}$$

$$\mathbf{A} = CNN_{\mathrm{AE}}(\mathbf{M}), \mathbf{A} \in \mathbb{R}^{h\times w\times n}, \tag{3}$$

$$\mathbf{F} = (\mathbf{V} \otimes \sigma(\mathbf{A}))/|hw|, \mathbf{F} \in \mathbb{R}^{c\times n}, \tag{4}$$

$$\mathbf{F}_{\mathrm{d}} = \{\mathbf{f}_1, \mathbf{f}_2, ..., \mathbf{f}_{n-1}\}, \mathbf{f} \in \mathbb{R}^{c}, \tag{5}$$

where $CNN_{\mathrm{VE}}$ and $CNN_{\mathrm{AE}}$ are value and attention extractors, respectively, i.e., a simple single convolution layer with a $3 \times 3$ kernel and $1 \times 1$ padding. $\mathbf{F}$ is the set of extracted feature vectors, $\mathbf{f}$ is a single vector in feature set $\mathbf{F}$, $\mathbf{F}_{\mathrm{d}}$ is the final feature set with the last discarded background feature, $|\cdot|$ is the matrix size, and $\sigma$ is softmax operation.

## 3.3 Channel-wise attention mechanism

A set of the weighted feature vectors $\mathbf{F}_{\mathrm{d}}$ with spatial attention are recalibrated by capturing and applying channel-wise attention, as depicted in the SE block that modulates channel activation (Hu *et al.*, 2020). Channel-wise attention adjusts the activation intensity according to the importance of each channel. This attention reduces the intensities of unnecessary feature elements, thereby reducing their influence on image distance calculations. Because each feature vector relates to a feature map, which is highlighted on a specific vehicle area, channel-wise attention should be applied to all $n-1$ feature vectors, which are in contrast to the original SE that controls one feature using FC layers.

We propose a channel-wise attention network based on an extended SE (ESE) algorithm. Furthermore, we re-shaped a set of the weighted feature vectors $\mathbf{F}_{\mathrm{d}}$ into one vector. We then feed the vec-tor to the SE block to modulate channel activation. The ESE module comprises two linear layers: the first and second layers are followed by a rectified linear unit and sigmoid operation, respectively. The ESE input dimension is 2048, and its output dimensions are 128 and 2048, in sequence. The result from the ESE is channel-wise attention, which is multiplied by the original $\mathbf{F}_d$. Thus, ESE can be summarized as follows:

$$\mathbf{f}_d = reshape(\mathbf{F}_d), \mathbf{f}_d \in \mathbb{R}^r, \quad (6)$$

$$\mathbf{f}_e = \mathbf{f}_d \times \rho(MLP(\mathbf{f}_d)), \mathbf{f}_e \in \mathbb{R}^r, \quad (7)$$

$$\mathbf{F}_e = reshape(\mathbf{f}_e), \mathbf{F}_e \in \mathbb{R}^{c\times(n-1)}, \quad (8)$$

where $r$ is $|c \times (n-1)|$, $MLP$ is a multi-layer perceptron, $\rho$ is the sigmoid operation, and $\mathbf{F}_e$ is a set of $n-1$ final recalibrated features. $\mathbf{f}_d$ and $\mathbf{f}_e$ are one-dimensional (1D) vectors, and $\mathbf{F}_d$ and $\mathbf{F}_e$ are 2D matrices.

## 3.4 Distance computation

The feature vector set ($\mathbf{F}_e$) extracted from the attention-based network and a feature vector ($\mathbf{f}_g$) obtained using the global average pooling of $\mathbf{M}$ are used to calculate losses. We separately apply triplet loss to each feature vector to train the proposed model, and we adopt a multi-feature re-weighting function called a co-occurrence attentive module (Chen *et al.*, 2020b), with some modifications, to calculate the feature distance by integrating these features for inference.

The weights of the distance between two vehicles are calculated as follows:

$$w_{(a,b),i} = \frac{AR_{a,i} \times AR_{b,i}}{\sum_{j=0}^{n-1} AR_{a,j} \times AR_{b,j}}, \quad (9)$$

where $AR_{a,\,i}$ is an area ratio with the i-th attention weight for the $a$-th image, and $AR$ is calculated by averaging the attention weights.

The original study (Chen *et al.*, 2020b) used weight = 1 for the global feature, whereas we use $\frac{1}{n-1}$ for the global feature weight $w_{(a,\,b),\,g}$. Hence, the distance between two vehicles is denoted as follows:

$$D_{(a,b)} = \sum_{i=0}^{n-1} w_{(a,b),i} \times \|\mathbf{f}_{a,i} - \mathbf{f}_{b,i}\|_2 + w_{(a,b),g} \times \|\mathbf{f}_{a,g} - \mathbf{f}_{b,g}\|_2, \quad (10)$$

where $\mathbf{f}_{a,\,i}$ is the $i$-th feature for the $a$-th image, $\mathbf{f}_{a,\,g}$ is a global feature, and $\|\cdot\|_2$ is the Euclidean distance.

# 4. Loss Function

We use three loss functions to train the model: cross-entropy loss for vehicle ID prediction ($\mathcal{L}_{id}$), triplet loss for distance learning ($\mathcal{L}_{tri}$), and production loss to separate each attention feature ($\mathcal{L}_{div}$). The overall loss function is given as follows:

$$\mathcal{L} = \mathcal{L}_{id} + \mathcal{L}_{tri} + \mathcal{L}_{div} \quad (11)$$

## 4.1 Cross-entropy loss

We apply cross-entropy loss following the vehicle ID prediction layer:

$$\mathcal{L}_{id} = -\frac{1}{K}\sum_{l=1}^{n}\sum_{i=1}^{K}\sum_{j=1}^{C} y_{i,j,l}\log(\hat{y}_{i,j,l}), \quad (12)$$

where $n$ is the number of features, $K$ is the number of images in a mini-batch, C is the number of classes, $y_{i,j,l}$ is the $j$th element for the one-shot encoded vector, which describes ground-truth for the $i$th sample in a mini-batch and $l$th feature vector, and $\hat{y}_{ijl}$ is the $j$th element of the output vector of the softmax FC layer for the $i$th image and $l$th feature vector.

## 4.2 Triplet loss

The proposed network is optimized using triplet loss for metric learning, which trains the network to minimize the distance between features from the same image classes and simultaneously maximizes the distance between features from different image classes. In a mini-batch that contains P identities and Q images for each identity, each image (anchor) has Q − 1 images of the same identity (positives) and $(P-1) \times Q$ images of different identities (negatives). Triplet loss (Hermans *et al.*, 2017) is defined as follows:

$$\mathcal{L}_{tri} = \sum_{e=1}^{n}\sum_{i=1}^{P}\sum_{a=1}^{Q}\left[m + \max_{\substack{p=1...Q\\ p\neq a}} D(\mathbf{v}_{a,i}, \mathbf{v}_{p,i}) - \min_{\substack{j=1...P\\ n=1...Q\\ j\neq i}} D(\mathbf{v}_{a,i}, \mathbf{v}_{n,j})\right]_+, \quad (13)$$

where $v_{a,\,i}$ is the prediction vector for the $a$-th image of the $i$-th identity group, and $m$ is the margin to control the difference between positive and negative pair distances, which helps cluster the distribution of the same vehicle images more densely.

## 4.3 Spatial diversity loss

We adopt spatial diversity loss (Chen *et al.*, 2020b) to restrict overlapped areas, hence ensuring that each attention weight acts on a different position:

$$\mathcal{L}_{div} = \sum_{i=1}^{K}(a_1^i \cdot a_2^i \cdot \ldots a_{n-1}^i), \quad (14)$$

where $a_n^i$ is $n$-th attention weight for the $i$-th image in the mini-batch, and spatial diversity loss is the summation of space-wise production of attention weights.

# 5. Experiments

## 5.1 Dataset

We tested the proposed method using three Re-ID datasets, such as VeRi-776, VehicleID, and VERI-Wild as follows:

(i) The VeRi-776 dataset (Liu *et al.*, 2016c) contains approximately 50 000 images, which comprises 776 vehicle ID images captured using 20 cameras. We used 576 IDs (37 778 images) for training and the remaining 200 vehicle IDs (11 579 images) for testing. The VeRi-776 data include vehicle colour and type labels.

(ii) The VehicleID dataset (Liu *et al.*, 2016a) contains 221 763 images from 26 267 vehicle IDs, which are arranged in three test sets (small, medium, and large) according to the number of query IDs (800, 1600, and 2400).

(iii) The VERI-Wild (Lou *et al.*, 2019b) is a recently created dataset from a more challenging environment than the other two. It contains 416 314 images of 40 671 vehicle IDs captured using 174 cameras over 1 month, under different weather and time conditions. Its dataset is the largest vehicle Re-ID dataset, which includes various weather environments in contrast to the previous two datasets. Similar to the VehicleID, the VERI-Wild also defines small, medium, and large test datasets with 3000, 5000, and 10 000 vehicle IDs, respectively.

**Table 2.** Model performance (mAP, CMC@1, and CMC@5) on the VehicleID.

| Method | Attention | Matadata | Small (800) | | | Medium (1600) | | | Large (2400) | | |
|---|---|---|---|---|---|---|---|---|---|---|---|
| | | | mAP | CMC@1 | CMC@5 | mAP | CMC@1 | CMC@5 | mAP | CMC@1 | CMC@5 |
| MD+CCL (Liu *et al.*, 2016c) | | ✓ | – | 0.490 | 0.735 | – | 0.428 | 0.668 | – | 0.382 | 0.616 |
| OIFE (Wang *et al.*, 2017) | | ✓ | – | – | – | – | – | – | – | 0.670 | 0.829 |
| RAM (Liu *et al.*, 2018) | | | – | 0.752 | 0.915 | – | 0.723 | 0.870 | – | 0.677 | 0.845 |
| EALN (Lou *et al.*, 2019a) | | ✓ | 0.775 | 0.751 | 0.881 | 0.742 | 0.718 | 0.839 | 0.710 | 0.693 | 0.814 |
| AAVER (Khorramshahi *et al.*, 2019a) | ✓ | ✓ | – | 0.747 | 0.938 | – | 0.686 | 0.900 | – | 0.635 | 0.856 |
| PRN (He *et al.*, 2019) | | ✓ | – | 0.784 | 0.923 | – | 0.750 | 0.883 | – | 0.742 | 0.864 |
| VAMI (Zhou & Shao, 2018) | ✓ | ✓ | – | 0.631 | 0.833 | – | 0.529 | 0.751 | – | 0.473 | 0.703 |
| SCAN (Teng *et al.*, 2018) | ✓ | | – | – | – | – | – | – | – | 0.654 | 0.785 |
| VCAM (Chen *et al.*, 2020a) | ✓ | ✓ | – | – | – | – | – | – | – | 0.661 | 0.761 |
| TAMR (Guo *et al.*, 2019) | ✓ | ✓ | 0.676 | 0.660 | 0.797 | 0.637 | 0.629 | 0.768 | 0.610 | 0.597 | 0.739 |
| UMTS (Jin *et al.*, 2020) | | | 0.870 | 0.809 | – | 0.842 | 0.788 | – | 0.828 | 0.761 | – |
| HPGN (Shen *et al.*, 2022) | | | 0.896 | 0.839 | – | 0.862 | 0.800 | – | 0.836 | 0.773 | – |
| SAVER (Khorramshahi *et al.*, 2020) | ✓ | | – | 0.799 | 0.952 | – | 0.776 | 0.911 | – | 0.753 | 0.883 |
| PVEN (Meng *et al.*, 2020) | | ✓ | – | 0.847 | 0.970 | – | 0.806 | 0.945 | – | 0.778 | 0.920 |
| PGAN (Zhang *et al.*, 2020) | ✓ | ✓ | – | – | – | – | – | – | – | 0.778 | 0.921 |
| MSA (Zheng *et al.*, 2020) | ✓ | | 0.803 | 0.776 | 0.905 | 0.771 | 0.744 | 0.863 | 0.756 | 0.729 | 0.844 |
| MVAN (Teng *et al.*, 2021) | ✓ | ✓ | – | – | – | – | – | – | – | 0.726 | 0.831 |
| PSA (Yang *et al.*, 2021) | ✓ | | – | 0.804 | 0.947 | – | 0.795 | 0.921 | – | 0.763 | 0.889 |
| SSIA (Li *et al.*, 2021) | ✓ | ✓ | – | <u>0.868</u> | 0.974 | – | <u>0.835</u> | <u>0.956</u> | – | 0.803 | <u>0.937</u> |
| CAL (Rao *et al.*, 2021) | ✓ | | 0.878 | 0.825 | 0.947 | 0.838 | 0.782 | 0.910 | 0.809 | 0.751 | 0.885 |
| Baseline | ✓ | | 0.872 | 0.82 | 0.965 | 0.853 | 0.794 | 0.934 | 0.831 | 0.776 | 0.905 |
| MUSP (This work) | ✓ | | **0.905** | **0.845** | <u>0.976</u> | **0.872** | **0.823** | **0.951** | **0.857** | <u>0.806</u> | **0.931** |

## 5.2 Implementation details

During pre-processing, we resized all images to 256 – 256 pixels and applied randomly erasing and translation effects. We used the Adam optimizer (Kingma & Ba, 2014) with weight decay and momentum of 5e−4 and 0.9, respectively. The proposed model was trained using a batch size of 64, 16 unique vehicle IDs, a training epoch of 90, and an initial learning rate of 0.000 35, which were divided by 10 at 30 and 60 epochs; we used the warm-up method with initial 10 epochs of 0.000 035–0.000 35. Furthermore, we applied label smoothing to avoid overfitting. Training required 6 and 2 h on the VehicleID and VeRi-776 datasets, respectively, and we used an NVIDIA Quadro RTX 6000 GPU system. The training code was written in PyTorch.

In the training phase, we used weighted feature vectors from the spatial attention module and vehicle ID prediction vector, as described in Section 4, for the loss function. However, we only used $\mathbf{F}_e$ and $\mathbf{f}_g$ in the inference phase with a re-weighting method to compute the distances between vehicles.

## 5.3 Baseline and compared methods

We set our baseline as a strong baseline using bag-of-tricks proposed by Luo *et al.* (2019). Our proposed model uses the attention modules to extract *n* weighted feature vectors. In contrast, the baseline replaces the attention modules with an average pooling layer; however, the remaining pre-processing process, learning process, and architecture are the same.

We compared our method with some SOTA methods. As selected and briefly described by Meng *et al.* (2020), BOW-CN (Zheng *et al.*, 2015) first adopted the BOW model based on the colour name (CN). Local maximal occurrence representation (LOMO; Liao *et al.*, 2015) is robust to the varied lighting conditions. The fusion of attributes and colour features (FACT; Liu *et al.*, 2016b) combines the low-level colour and high-level semantic features. The BOW-CN, LOMO, and FACT are handcraft feature-based methods. The remaining are deep learning-based methods. GoogLeNet (Yang *et al.*, 2015a) is a GoogleNet (Szegedy *et al.*, 2015) model fine-tuned on the CompCars (Yang *et al.*, 2015b) dataset. Plate-SNN (Liu *et al.*, 2018) uses the number-plate features to enhance the retrieval vehicles. Siamese+Path (Shen *et al.*, 2017) was used to propose the visual-spatial-temporal path to exploit the temporal restriction. GSTE (Bai *et al.*, 2018) proposed group-sensitive-triplet embedding to elegantly model the intra-class variance. VAMI (Zhou & Shao, 2017) generated features of different views using GAN, while feature distance adversarial network (FDA-Net; Lou *et al.*, 2019b) generated the hard negative samples in the feature space. EALN (Lou *et al.*, 2019c) proposed an adversarial network that can generate samples localized in the embedding space. OIFE (Wang *et al.*, 2017) used the 20 pre-defined key points to roughly align vehicle features. RAM (Liu *et al.*, 2018) horizontally split the image into three parts. PRN (He *et al.*, 2019) detected the window, light, and brand of a vehicle to capture the difference between vehicle instances. AAVER (Khorramshahi *et al.*, 2019a) proposed an attention mechanism based on vehicle key points and orientation. UMTS (Jin *et al.*, 2020) introduced an uncertainty-aware multi-shot network composed of teacher and student networks. HPGN (Shen *et al.*, 2022) used a hybrid pyramidal graph network to explore the spatial significance of feature maps. We added the following recent representative attention-based deep learning methods, as discussed in Section 2.2: SCAN (Teng *et al.*, 2018), VCAM (Chen *et al.*, 2020a), SPAN (Chen *et al.*, 2020b), SAVER (Khorramshahi *et al.*, 2020), PVEN (Meng *et al.*, 2020), PGAN (Zhang *et al.*, 2020), MSA (Zheng *et al.*, 2020), MVAN (Teng *et al.*, 2021), PSA (Yang *et al.*, 2021), SSIA (Li *et al.*, 2021), CAL (Rao *et al.*, 2021), and MUSP.

## 5.4 Experiments on the VehicleID dataset

Table 2 compares VehicleID dataset outcomes for the baseline, MUSP, and existing attention-based vehicle Re-ID models using CMC@1 and CMC@5 metrics. The highest value without additional metadata is shown in bold, whereas the highest value is underlined in italics. The performance of the MUSP improved for all CMC metrics compared with the baseline, with 2.5% and 1.1%, 2.9% and 1.7%, and 3.0% and 2.6% improvement on CMC@1 and CMC@5 for the small, medium, and large test sets, respectively. Compared with the existing methods, MUSP showed the highest value in two CMC cases, whereas SSIA (Li *et al.*, 2021) recorded the highest value in four of six CMC cases. However, SSIA is assisted by additional metadata and self-supervised learning. Among the non-metadata models, MUSP recorded the highest performance in all metrics for three VehicleID datasets. Therefore, MUSP can outperform the other methods if additional metainformation is used.

**Table 3.** Model performance (mAP, CMC@1, and CMC@5) on the VERI-Wild.

| Method | Attention | Matadata | Small (3000) | | | Medium (5000) | | | Large (10 000) | | |
|---|---|---|---|---|---|---|---|---|---|---|---|
| | | | mAP | CMC@1 | CMC@5 | mAP | CMC@1 | CMC@5 | mAP | CMC@1 | CMC@5 |
| GoogLeNet (Yang *et al.*, 2015a) | | | 0.243 | 0.572 | 0.751 | 0.242 | 0.532 | 0.711 | 0.215 | 0.446 | 0.636 |
| Triplet (Schroff *et al.*, 2015) | | | 0. 157 | 0.447 | 0.633 | 0. 133 | 0.403 | 0.590 | 0. 099 | 0.335 | 0. 514 |
| Softmax (Liu *et al.*, 2017) | | | 0.264 | 0.534 | 0. 750 | 0.227 | 0.462 | 0. 699 | 0.176 | 0.379 | 0. 599 |
| CCL (Liu *et al.*, 2016a) | | | 0.225 | 0.570 | 0. 750 | 0.193 | 0.519 | 0. 710 | 0.148 | 0.446 | 0.610 |
| HDC (Yuan *et al.*, 2017) | | | 0. 291 | 0.571 | 0.789 | 0. 248 | 0.496 | 0.723 | 0. 183 | 0.440 | 0.649 |
| GSTE (Bai *et al.*, 2018) | | | 0.314 | 0.605 | 0.801 | 0.262 | 0.521 | 0.749 | 0.195 | 0.454 | 0.665 |
| Unlabelled GAN (Zhu *et al.*, 2017) | | | 0.299 | 0.581 | 0.796 | 0.247 | 0.516 | 0.744 | 0.182 | 0.436 | 0.655 |
| FDA-Net (Lou *et al.*, 2019b) | | | 0.351 | 0.640 | 0.828 | 0.298 | 0.578 | 0.783 | 0.228 | 0.494 | 0.705 |
| UMTS (Jin *et al.*, 2020) | | | 0.828 | 0.845 | – | 0.661 | 0.793 | – | 0.542 | 0.728 | – |
| HPGN (Shen *et al.*, 2022) | | | 0.804 | 0.913 | – | 0.751 | 0.882 | – | 0.650 | 0.826 | – |
| SAVER (Khorramshahi *et al.*, 2020) | ✓ | | 0.809 | 0.945 | 0.981 | 0.753 | 0.927 | 0.974 | 0.677 | 0.895 | 0.958 |
| PVEN (Meng *et al.*, 2020) | ✓ | ✓ | 0. 825 | 0.967 | 0.992 | 0.770 | 0.954 | 0.988 | 0.697 | 0.934 | 0.978 |
| Baseline | ✓ | | 0.798 | 0.952 | 0.987 | 0.740 | 0.935 | 0.982 | 0.660 | 0.909 | 0.969 |
| MUSP (This work) | ✓ | | 0.846 | **0.961** | **0.989** | 0.796 | **0.947** | **0.987** | 0.726 | **0.927** | **0.977** |

**Table 4.** Model performance (mAP, CMC@1, and CMC@5) on the VeRi-776. A and M denote using attentions and using additional metadata, respectively.

| Method | A | M | mAP | CMC@1 | CMC@5 |
|---|---|---|---|---|---|
| BOW-CN (Zheng *et al.*, 2015) | | | 0.122 | 0.339 | 0.537 |
| LOMO (Liao *et al.*, 2015) | | | 0.096 | 0.253 | 0.465 |
| GoogLeNet (Yang *et al.*, 2015a) | | | 0.170 | 0.498 | 0.712 |
| FACT (Liu *et al.*, 2016b) | | | 0.185 | 0.510 | 0.735 |
| Plate-SNN (Liu *et al.*, 2018) | | | 0.278 | 0.614 | 0.788 |
| Siamese+Path (Shen *et al.*, 2017) | | | 0.583 | 0.835 | 0.900 |
| OIFE (Wang *et al.*, 2017) | | ✓ | 0.480 | 0.894 | – |
| RAM (Liu *et al.*, 2018) | | | 0.615 | 0.886 | 0.940 |
| EALN (Lou *et al.*, 2019c) | | ✓ | 0.574 | 0.844 | 0.941 |
| AAVER (Khorramshahi *et al.*, 2019a) | ✓ | ✓ | 0.612 | 0.890 | 0.947 |
| PRN (He *et al.*, 2019) | | ✓ | 0.743 | 0.943 | 0.989 |
| VAMI (Zhou & Shao, 2017) | ✓ | ✓ | 0.501 | – | – |
| SCAN (Teng *et al.*, 2018) | ✓ | | 0.499 | 0.822 | 0.908 |
| VCAM (Chen *et al.*, 2020a) | ✓ | ✓ | 0.686 | 0.944 | 0.969 |
| SPAN (Chen *et al.*, 2020b) | ✓ | ✓ | 0.689 | 0.940 | 0.976 |
| UMTS (Jin *et al.*, 2020) | | | 0.759 | 0.958 | – |
| HPGN (Shen *et al.*, 2022) | | | 0.802 | 0.967 | – |
| SAVER (Khorramshahi *et al.*, 2020) | ✓ | | 0.796 | 0.964 | 0.986 |
| PVEN (Meng *et al.*, 2020) | ✓ | ✓ | 0.795 | 0.956 | 0.984 |
| PGAN (Zhang *et al.*, 2020) | ✓ | ✓ | 0.793 | 0.965 | 0.983 |
| MSA (Zheng *et al.*, 2020) | ✓ | | 0.629 | 0.921 | 0.962 |
| MVAN (Teng *et al.*, 2021) | ✓ | ✓ | 0.725 | 0.926 | 0.979 |
| PSA (Yang *et al.*, 2021) | ✓ | | 0.759 | 0.939 | 0.963 |
| SSIA (Li *et al.*, 2021) | ✓ | ✓ | 0.810 | 0.967 | 0.986 |
| CAL (Rao *et al.*, 2021) | ✓ | | 0.743 | 0.954 | 0.979 |
| Baseline | ✓ | | 0.768 | 0.952 | 0.976 |
| MUSP (This work) | ✓ | | **0.780** | **0.956** | **0.979** |

**Table 5.** MUSP performance on the VeRi-776 using the number of attentions.

| Number of attentions | mAP | CMC@1 | CMC@5 |
|---|---|---|---|
| 6 | 0.771 | 0.951 | 0.976 |
| 5 | **0.780** | **0.956** | **0.979** |
| 4 | 0.777 | **0.956** | 0.978 |
| 3 | 0.763 | 0.953 | 0.977 |
| 2 | 0.769 | 0.951 | 0.972 |

## 5.5 Experiments on the VERI-Wild dataset

Table 3 compares the performance of the baseline, MUSP, and various existing vehicle Re-ID models using mAP, CMC@1, and CMC@5. For the mAP metric, MUSP exhibits a remarkable performance improvement compared with the baseline, achieving 4.8%, 5.6%, and 6.6% improvements for the small, medium, and large datasets, respectively. The performance improvement is particularly noticeable for complex test sets with numerous IDs. The MUSP achieved a SOTA performance with 2.1%, 2.6%, and 2.9% improvements over PVEN, which is the current SOTA method for the small, medium, and large datasets, respectively. Compared with the baseline and PVEN, the MUSP improved as the test dataset size increased. Compared with the baseline, the MUSP improved by 0.9%, 1.2%, and 1.8% on the CMC@1 for the small, medium, and large datasets, respectively. For CMC@1 and CMC@5, the MUSP achieve SOTA among the non-metadata models and comparable performance to PVEN (Meng *et al.*, 2020), which is the current SOTA that uses metadata. Thus, MUSP consistently improves performance across all test sets, confirming that increasing model representation capability is sufficient for achieving comparable or superior performance.

## 5.6 Experiments on the VeRi-776 dataset

Table 4 compares the VeRi-776 dataset outcomes for the baseline, MUSP, and previous vehicle Re-ID models using mAP, CMC@1, and CMC@5 metrics. Compared with the baseline, MUSP improved by 1.2% and 0.4% for mAP and CMC@1, respectively. MUSP shows comparable performance compared with the models that use additional metadata (e.g., AAVER, PRN, and PVEN). However, it does not require metadata and achieves a reasonable performance improvement using only different attention mechanisms. The SSIA (Li *et al.*, 2021) achieved SOTA for all metrics. However, as mentioned above, SSIA is assisted by additional metainforma-tion and self-supervised learning. Among the non-metadata models, MUSP recorded the highest performance in all three metrics.

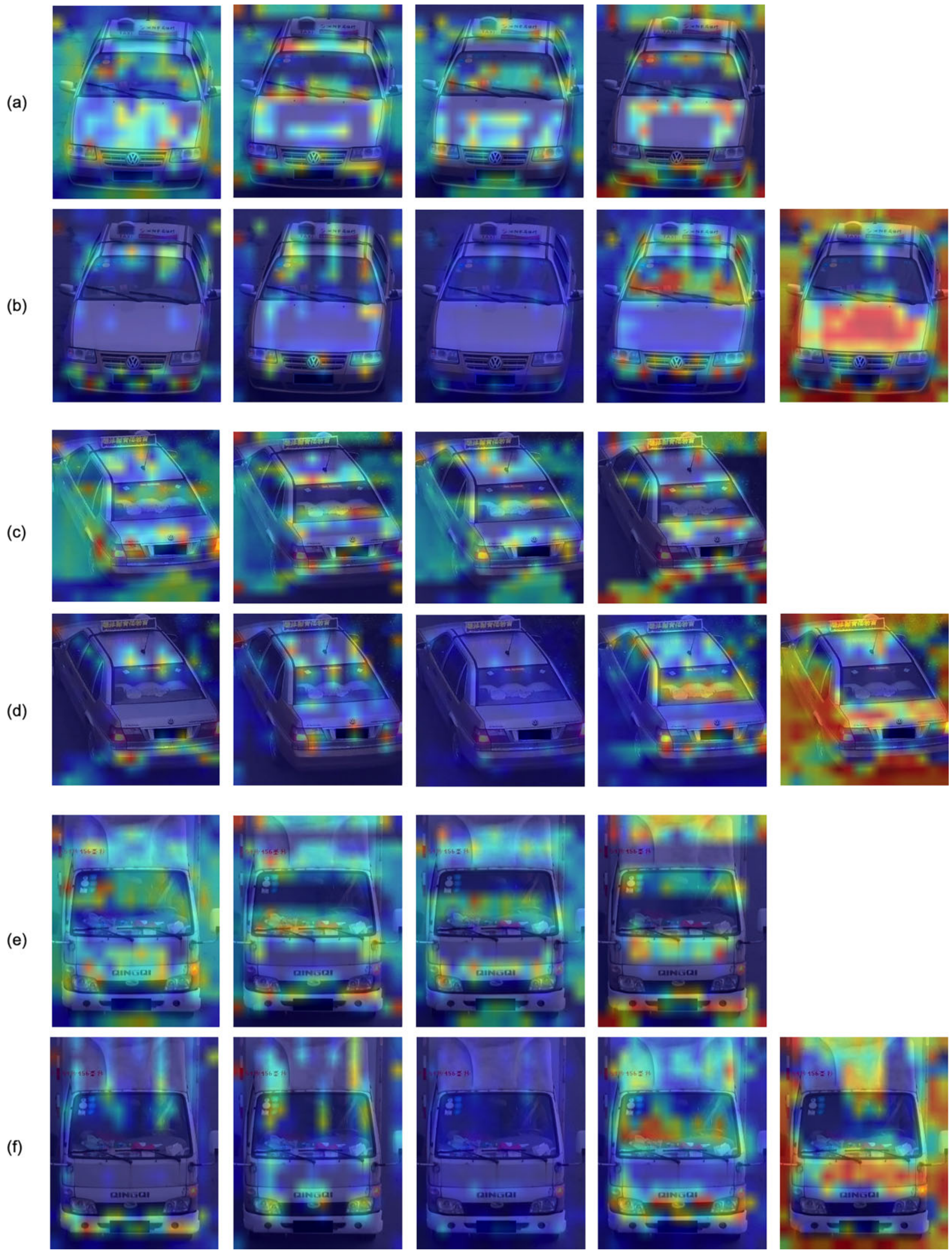

**Figure 4:** The effect of omitting a spatial attention map from the pipeline: (a), (c), and (e) are heat maps depicting the attention weights of each map when all four spatial attention maps are used. (b), (d), and (f) are heat maps depicting the attention weights of each map when the last spatial attention map among the five is not used in subsequent computation for the final result.

**Table 6.** Improvement in model performance when omitting a spatial attention map from the pipeline. For the Veri-776 and VehicleID datasets, the performance of the following two models was compared. In the first model, one of the five spatial attention maps is excluded from the pipeline, whereas in the second model, all four spatial attention maps are used for training and inference. The result showed that the first model outperformed the second model.

| Method | VeRi-776 | | | VehicleID | | | | | | | | |
|---|---|---|---|---|---|---|---|---|---|---|---|---|
| | | | | Small (800) | | | Medium (1600) | | | Large (2400) | | |
| | mAP | CMC@1 | CMC@5 | mAP | CMC@1 | CMC@5 | mAP | CMC@1 | CMC@5 | mAP | CMC@1 | CMC@5 |
| 4 Attentions | 0.767 | 0.947 | 0.977 | 0.898 | 0.842 | 0.974 | **0.876** | 0.820 | 0.951 | 0.857 | 0.803 | 0.931 |
| 5-1 Attentions | **0.780** | **0.956** | **0.979** | **0.905** | **0.845** | **0.976** | 0.872 | **0.823** | 0.951 | 0.857 | **0.806** | 0.931 |

**Table 7.** The effects of the channel-wise attention module on the VehicleID dataset (small test). The ESE is the main part of the module.

| Method | CMC@1 | CMC@5 |
|---|---|---|
| MUSP w ESE | **0.845** | **0.976** |
| MUSP w/o ESE | 0.830 | 0.961 |

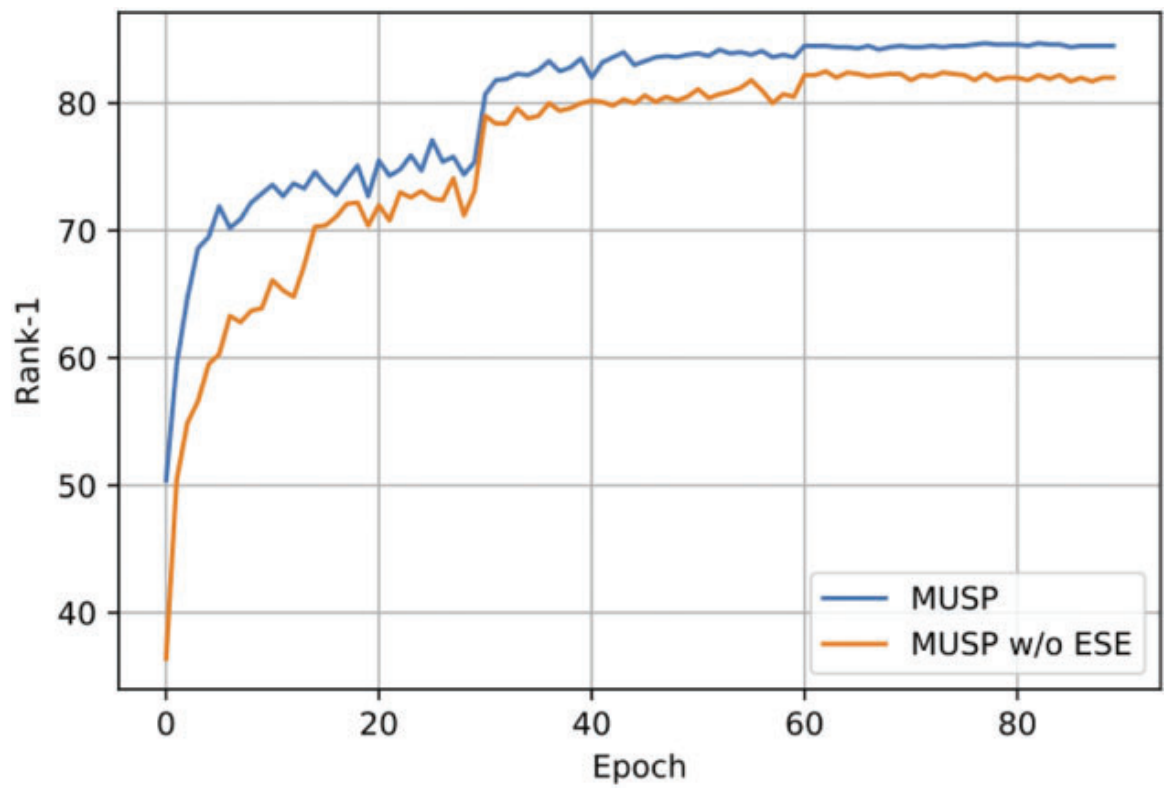

**Figure 5:** ESE impact on the MUSP performance using training epoch on the small VehicleID dataset. The MUSP achieves consistently higher performance with faster convergence when including the ESE module.

**Table 8.** Sigmoid- and softmax-based attention modules for the MUSP on the VeRi-776.

| Method | mAP | CMC@1 | CMC@5 |
|---|---|---|---|
| MUSP w softmax | **0.780** | **0.956** | **0.979** |
| MUSP w sigmoid | 0.772 | 0.954 | 0.977 |

**Table 9.** The mAP, CMC@1, and CMC@5 on cross-domain settings.

| Method | Train | Test | CMC@1 | CMC@5 |
|---|---|---|---|---|
| RAM (Liu *et al.*, 2018) | VehicleID | VehicleID | 0.752 | 0.915 |
| EALN (Lou *et al.*, 2019a) | VehicleID | VehicleID | 0.751 | 0.881 |
| PVEN (Meng *et al.*, 2020) | VERI-Wild | VehicleID | 0.772 | 0.944 |
| Baseline | VERI-Wild | VehicleID | 0.737 | 0.931 |
| MUSP (This work) | VERI-Wild | VehicleID | **0.797** | **0.951** |

## 5.7 Ablation study

### 5.7.1 *Number of spatial attention maps*

According to Chen *et al.* (2020a)'s interpretation, the spatial attention mechanism extracts viewpoint-aware features and the channel-wise attention mechanism detects discriminative vehicle parts. Experiments were conducted to observe how the number of attentions affects performance and the number of attention optimal for the experimental datasets. Generally, increasing the number of attention increases the number of views, thereby improving model performance. Reducing the number will have the opposite effect. Table 5 presents the optimal performance for the five attention. Those less than three produced a significantly lower performance than four or more attention. Therefore, it can be interpreted that few viewpoint-aware attention cannot capture the discriminative features of a vehicle from sufficiently diverse angles. The experiments illustrate that the desired part recognition and comparison can only be performed if four or more spatial attention are used.

### 5.7.2 *Effects of omitting a spatial attention map*

Figure 4 shows the heat maps corresponding to the weights of spatial attention maps trained using the VehicleID dataset. Figures 4(a), (c), and (e) illustrate the attention distribution when all four spatial attention maps are used for inference. Figures 4(b), (d), and (f) show the attention distribution when four out of five attention maps are used. As shown in the figures, excluding one attention map from the computational pipeline during training considerably reduces the noise. We also compared the performance of the two models for the VeRi-776 and VehicleID datasets. As shown in Table 6, the model using four out of five attention maps outperforms the model using all four spatial attention maps for training and inference.

### 5.7.3 *Effects of the channel-wise attention module*

Table 7 compares the MUSP performance with and without the channel-wise attention module on the VehicleID dataset. The ESE is the central element of the channel-wise attention module. It improved by 0.015 for the CMC@1 and CMC@5, verifying that it successfully recalibrated channel information, which is helpful for distance computation and embedding. It receives $n-1$ features representing discriminative vehicle part regions from an arbitrary viewpoint and adjusts the channel activation by considering the overall system performance. The channel activation is reduced using ESE to suppress its effect when some channel information is unnecessary for vehicle comparison. Figure 5 compares the ESE impact using epoch. Here, the ESE improved performance for all epochs and accelerated convergence.

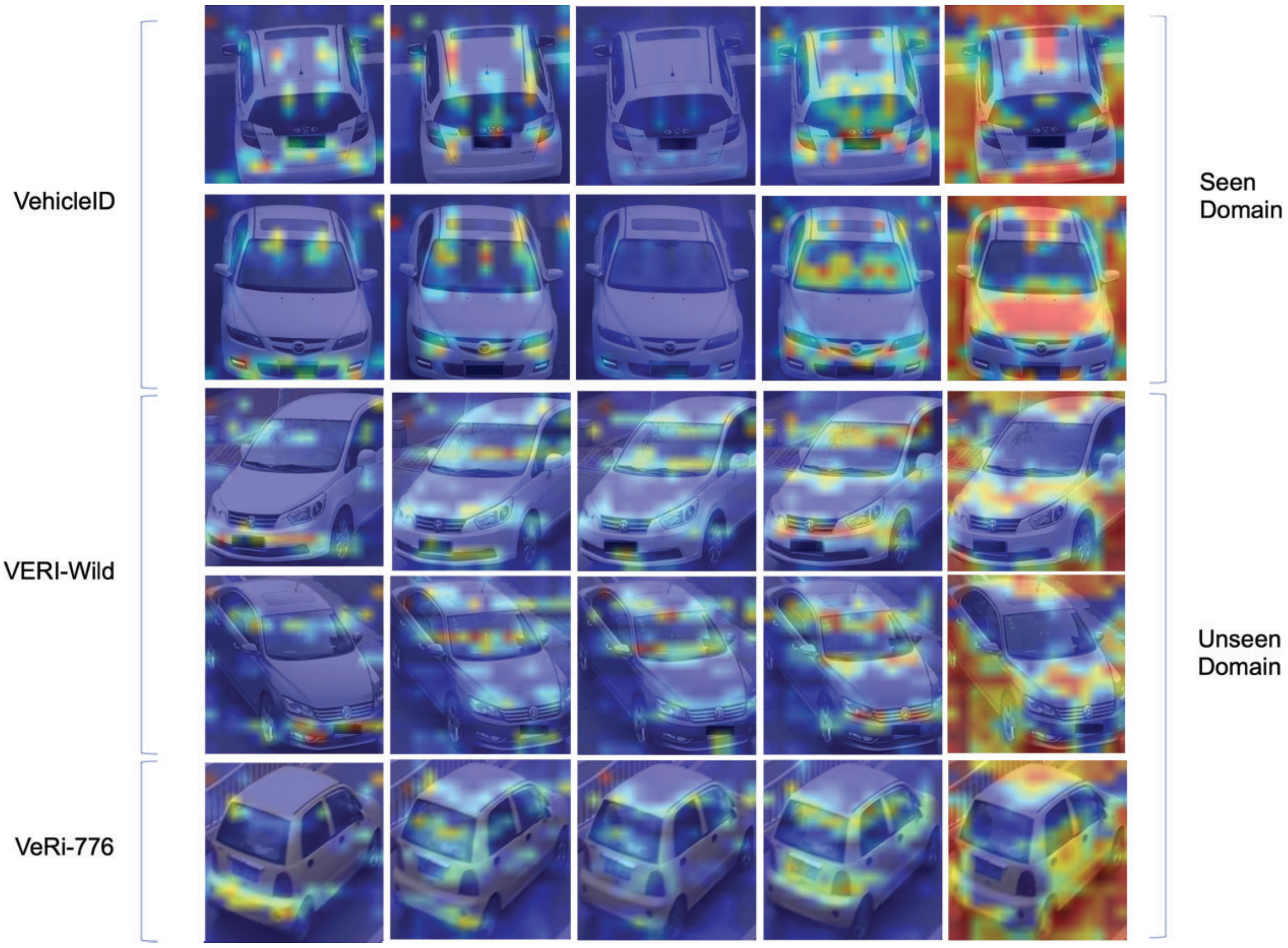

**Figure 6:** Results of applying the MUSP to three primary vehicle Re-ID datasets. This model is trained using the VehicleID dataset and tested on three different datasets. Each column corresponds to the attention maps. Regardless of orientation and vehicle type, each attention has a high activation on one or more specific regions. For example, the first attention reacts to the bottom of the headlights, the second reacts to the headlights, and the third reacts to the front or rear window. The fourth has a high activation on the lower parts of the window and vehicle and the fifth is for the background. Attention is invariant to the domain and orientation attributes even for the unseen domains not used in training the model (e.g., the VERI-Wild and VeRi-776 datasets).

#### 5.7.4 *Activation functions of spatial attention module*

CBAM (Woo *et al.*, 2018) and SENet (Hu *et al.*, 2020) used a sigmoid-based attention module, whereas the proposed spatial attention module is based on softmax. Softmax satisfies our spatial partition purposes more closely because it has a normalization effect that sets the sum of the dimension elements to one. Combining softmax and spatial diversity loss produces exclusively spatial activation. Gradient vanishing can occur for the sigmoid approach as training progresses, thus degrading performance. We compared the softmax- and sigmoid-based attention modules to verify that the former is the more suitable activation function. The spatial attention module discards the last attention weight, so we retained four and five attention for the sigmoid- and softmax-based modules, respectively. Table 8 illustrates that the softmax-based attention improved by 0.8%, 0.3%, and 0.2% compared with the sigmoid-based attention for the mAP, CMC@1, and CMC@5 metrics, respectively. Thus, it showed superior overall performance improvement to the sigmoid-based attention.

### 5.8 Cross-domain experiments

These experiments confirm that the proposed MUSP performs particularly well for larger test datasets close to real-world environments. Recognizing previously unseen vehicles is another problem that emerges in real-world environments. Therefore, we conducted a cross-domain experiment comparing the RAM (Lin *et al.*, 2014) and EALN (Lou *et al.*, 2019a), which are trained and tested on the VehicleID, with the PVEN (Meng *et al.*, 2020) and MUSP trained on the VERI-Wild and tested on the VehicleID.

Table 9 presents an overwhelming MUSP performance for the cross-domain tests; the attention partition operates effectively even for vehicles not previously learned. The MUSP exceeded the models trained on the same dataset and improved by 6% and 2.5% compared with the baseline and PVEN, respectively. These results are consistent with those in Fig. 6, where vehicle parts on the VeRi-776 and VERI-Wild datasets were equally identified, although the MUSP was trained on the VehicleID dataset.

## 6. Conclusions

In this study, we proposed a MUSP network based on multiple soft attentions for vehicle Re-ID. This model can learn to highlight the most discriminant regions and suppress the distraction of irrelevant parts using multiple spatial attention and channel-wise attention mechanisms without introducing any artificial part-specific and view-dependent attention branches. We introduced a novel method that can considerably reduce noise in spatial attention maps, thereby solving the problem of noise generation by spatial attention mechanisms in attention maps. This architecture achieved significant performance improvement without using additional metadata. The visualization in Fig. 6 illustrates that the attention parts selected using the spatial attention module also operate effectively on unseen data and perform invariant orientations. The spatial and channel-wise attention modules are vital MUSP components, which were experimentally verified on the VehicleID, VERI-Wild, and VeRi-776 datasets. The experiments showed that the proposed method was comparable to or superior to current SOTA methods. Particularly, among the attention-based models that do not use additional metadata, our model achieved a SOTA and showed comparable performance to the methods that used metadata. As this performance is achieved without using metadata attributes or post-processing, such as re-ranking, the performance of the model can be significantly improved when they are used.

For future research, we will consider applying the MUSP to the feature maps extracted from various levels of the layers of a backbone network, such as the SENet. The MUSP is currently applied only to the resulting feature map of the last layer of the backbone. However, applying it to feature maps extracted from the middle layers of the backbone will significantly improve performance. Furthermore, the proposed multiple soft attention method can be applied to various areas (Mohammed *et al.*, 2022; Shi *et al.*, 2022; Yan, 2018), such as monitoring traffic (Baek & Lee, 2022; Jain *et al.*, 2019), production lines (Park & Lee, 2022), or intelligent human-machine systems (Eom & Lee, 2015, 2022; Lee & Yoon, 2020).

## Acknowledgments

This work was supported by the National Research Foundation of Korea (NRF) grant funded by the Korea government (MSIT) (No. 2020R1A2C1102767). This is an extended and revised version of a preliminary conference paper that was presented in the 25th IEEE International Conference on Intelligent Transportation Systems (ITSC 2022) (Lee *et al.*, 2022).

## Conflict of interest statement

None declared.